\documentclass{article}
\usepackage{spconf}
\usepackage[utf8]{inputenc} 
\usepackage[T1]{fontenc}    
\usepackage{hyperref}       
\usepackage{url}            
\usepackage{booktabs}       
\usepackage{amsfonts}       
\usepackage{amsmath}
\usepackage{amssymb}
\usepackage{caption}
\usepackage{subcaption}
\usepackage{multirow}
\usepackage{graphicx}
\usepackage{nicefrac}       
\usepackage{microtype}      
\usepackage{xcolor}         
\usepackage{csquotes}
\usepackage{makecell}
\usepackage{tipa}

\title{Multi-Task Learning for Front-End Text Processing in TTS}
\name{Wonjune Kang\textsuperscript{\textnormal{1}*\thanks{*Work done as an intern at Meta.}}, Yun Wang\textsuperscript{\textnormal{2}}, Shun Zhang\textsuperscript{\textnormal{2}}, Arthur Hinsvark\textsuperscript{\textnormal{2}}, Qing He\textsuperscript{\textnormal{2}}}
\address{\textsuperscript{1}Massachusetts Institute of Technology \quad \textsuperscript{2}AI at Meta}

\begin{document}
\newcommand\mdoubleplus{\ensuremath{\mathbin{+\mkern-10mu+}}}
\ninept
\maketitle
\begin{abstract}
We propose a multi-task learning (MTL) model for jointly performing three tasks that are commonly solved in a text-to-speech (TTS) front-end: text normalization (TN), part-of-speech (POS) tagging, and homograph disambiguation (HD).
Our framework utilizes a tree-like structure with a trunk that learns shared representations, followed by separate task-specific heads.
We further incorporate a pre-trained language model to utilize its built-in lexical and contextual knowledge, and study how to best use its embeddings so as to most effectively benefit our multi-task model.
Through task-wise ablations, we show that our full model trained on all three tasks achieves the strongest overall performance compared to models trained on individual or sub-combinations of tasks, confirming the advantages of our MTL framework.
Finally, we introduce a new HD dataset containing a balanced number of sentences in diverse contexts for a variety of homographs and their pronunciations.
We demonstrate that incorporating this dataset into training significantly improves HD performance over only using a commonly used, but imbalanced, pre-existing dataset.
\end{abstract}

\vspace{-5.5pt}
\section{Introduction}
\label{sec:introduction}
\vspace{-5.5pt}

The front-end of a text-to-speech (TTS) pipeline plays an essential role in the performance of the overall system, taking on a variety of linguistic tasks that convert input text into phonetic representations.
While the exact components of a TTS front-end can vary, some tasks that are commonly addressed include text normalization (TN)~\cite{sproat2001normalization}, part-of-speech (POS) tagging~\cite{pouget2016adaptive}, and homograph disambiguation (HD)~\cite{yarowsky1997homograph}, leading up to grapheme-to-phoneme conversion (G2P)~\cite{bisani2008joint}.
In recent years, data-driven approaches utilizing deep neural networks have seen great success in these tasks, notably for TN~\cite{sproat2016rnn, zhang2019neural, mansfield2019neural}, HD~\cite{gorman2018improving, nicolis2021homograph}, and as end-to-end front-ends~\cite{conkie2020scalable}.

In this work, we consider how to better solve TN, POS tagging, and HD in the context of a TTS front-end for American English.
Although all three tasks share a common input (the text to be synthesized into speech), in most pipelines, the modules that solve each task are usually trained and used separately.
Intuitively, however, one might expect them to be able to take advantage of shared representations containing common high-level information.
For example, POS information could help with recognizing non-standard words such as numbers or abbreviations in TN, or with determining the pronunciation of a word given a sentence's context in HD.
Additionally, certain cases in TN can be treated similarly to a homograph or word sense disambiguation problem (e.g., ``St. Mary's St.'' $\rightarrow$ ``Saint Mary's Street'').
This makes the three tasks opportune targets for multi-task learning (MTL), which can allow a model to capture more generalized and complementary knowledge that benefits its performance~\cite{caruana1997multitask}.

We propose a multi-task learning model for TN, POS tagging, and HD that aims to capitalize on the above-mentioned commonalities between the three tasks.
Our model has a tree-like structure with a shared trunk for general feature extraction and task-specific heads.
The trunk consists of two information streams that are combined using a cross-attention mechanism: the first operates on a token sequence for TN (described in Section~\ref{subsec:preliminaries}), and the second utilizes an embedding sequence from a pre-trained language model (LM)~\cite{lan2019albert}.
We investigate which layers of the LM to extract embeddings from and how to best incorporate them into the model so as to optimally benefit each task.
We also perform task-wise ablations to study how jointly learning different combinations of the three tasks affects model performance, and in doing so, we justify our intuition for the MTL framework by validating the presence of inter-task positive transfer.
Finally, we address a key gap in the HD literature: the lack of a strong dataset with balanced samples for different homograph pronunciations.
We introduce a new dataset that expands upon a commonly used, but imbalanced pre-existing corpus~\cite{gorman2018improving}; our dataset contains an equal number of sentences using each homograph's pronunciation in diverse contexts, generated using Llama 2~\cite{touvron2023llama2}.
We demonstrate that incorporating this dataset into training significantly improves HD performance over using only the imbalanced pre-existing dataset.\footnote{The dataset is publicly available at: \url{https://github.com/facebookresearch/llama-hd-dataset}}

In summary, the contributions of this paper are as follows: 1) We introduce a multi-task learning model for TN, POS tagging, and HD, and propose various architectural design choices to optimize its performance. 2) We justify the intuition behind our MTL approach via task-wise ablation studies that demonstrate the presence of positive transfer between the three tasks. 3) We introduce a new dataset for HD that extends upon the dataset from \cite{gorman2018improving} with balanced samples for all homograph pronunciations, and show that incorporating it into training significantly improves performance on the task.
\vspace{-6pt}
\section{Background and Related Work}
\label{sec:background}
\vspace{-6pt}

\subsection{TTS front-end tasks}
\vspace{-5pt}

\noindent \textbf{Text normalization.}
In the context of TTS, text normalization (TN) is the task of converting written text into its spoken form, transforming non-standard words into their appropriate verbalizations given the sentence's context.
Such non-standard words can be further categorized into semiotic classes~\cite{esch2017expanded} such as numbers, dates, or currency.

Traditional methods for TN used hand-crafted rules or handwritten grammars to verbalize input tokens~\cite{sproat1996multilingual, roark2012opengrm}.
More recent deep learning approaches have seen success in treating TN as a sequence-to-sequence problem~\cite{sproat2016rnn, sproat2017rnn, ro2022transformer}; however, these methods are susceptible to ``unrecoverable'' errors that can fundamentally change the meaning of an utterance (e.g. ``7/8 inches'' $\rightarrow$ ``five eighth inches'')~\cite{zhang2019neural}.
Other approaches have cast TN as a \textit{semiotic classification} task~\cite{tyagi2021proteno}.
Here, the procedure is to predict a class for each input token and perform normalization according to predetermined mechanisms associated with the class.
Because there is a limited set of known transformations that can be applied to each class, these methods provide more deterministic safeguards against unrecoverable errors.

\vspace{2pt}
\noindent \textbf{Part-of-speech tagging.}
Part-of-speech (POS) tagging has direct links to other tasks that are often part of a TTS front-end, such as homograph disambiguation~\cite{yarowsky1997homograph}.
While traditional approaches used hand-crafted rules or statistical methods~\cite{collins2002discriminative}, more recent ones have used neural models to achieve state-of-the-art performance, often using contextual word representations~\cite{elkahky2018challenge, bohnet2018morphosyntactic}.
Notably, POS tagging has been shown to be a near-universal helper task for text-based MTL models~\cite{changpinyo2018multi}, motivating its inclusion in our framework.

\vspace{2pt}
\noindent \textbf{Homograph disambiguation.}
Homograph disambiguation (HD) is often done before the G2P module in a TTS front-end to determine which pronunciation of a homograph to use.
It was traditionally done using rule-based or statistical decision procedures that utilized syntactic patterns~\cite{yarowsky1997homograph}.
More recently, \cite{gorman2018improving} proposed a supervised multinomial log-linear model that uses word context, POS tag, and capitalization features, and \cite{nicolis2021homograph} utilized contextual word embeddings from pre-trained Transformer language models as inputs to homograph-wise pronunciation classifiers.
We take inspiration from their findings in the design of our multi-task model.

\vspace{-9pt}
\subsection{Multi-task learning for text data}
\vspace{-5pt}

Many works have explored the applicability of multi-task learning (MTL) to text-based tasks~\cite{collobert2008unified, liu2019multi}.
The idea behind MTL is that jointly learning to solve multiple related tasks can allow a model to learn common knowledge that can benefit all of them, leading to more robustness and generalizability.
The concepts of \textit{positive} and \textit{negative transfer} play an important role here; that is, whether jointly learning pairs of tasks results in better or worse performance compared to separately learning each task, respectively~\cite{wu2019understanding}.

Several previous works incorporated TN, POS tagging, and/or HD-like tasks in MTL frameworks. \cite{higashiyama2021text} studied joint word segmentation, POS tagging, and lexical normalization in Japanese. \cite{li2015joint} proposed a joint model for POS tagging and TN on social media data, but their TN task involved converting non-standard language used online to standard form rather than written to spoken form.
Recently, \cite{ying2023unified} introduced a unified English TTS front-end that performs TN, prosody prediction, and G2P, with POS tagging and HD as intermediate steps.
However, it did not provide an in-depth analysis of how jointly learning the different tasks affects performance on each one.
To the best of our knowledge, no previous works have studied the concrete impact of multi-task learning on various TTS front-end tasks.
\begin{figure}
    \centering
    \includegraphics[width=0.88\columnwidth]{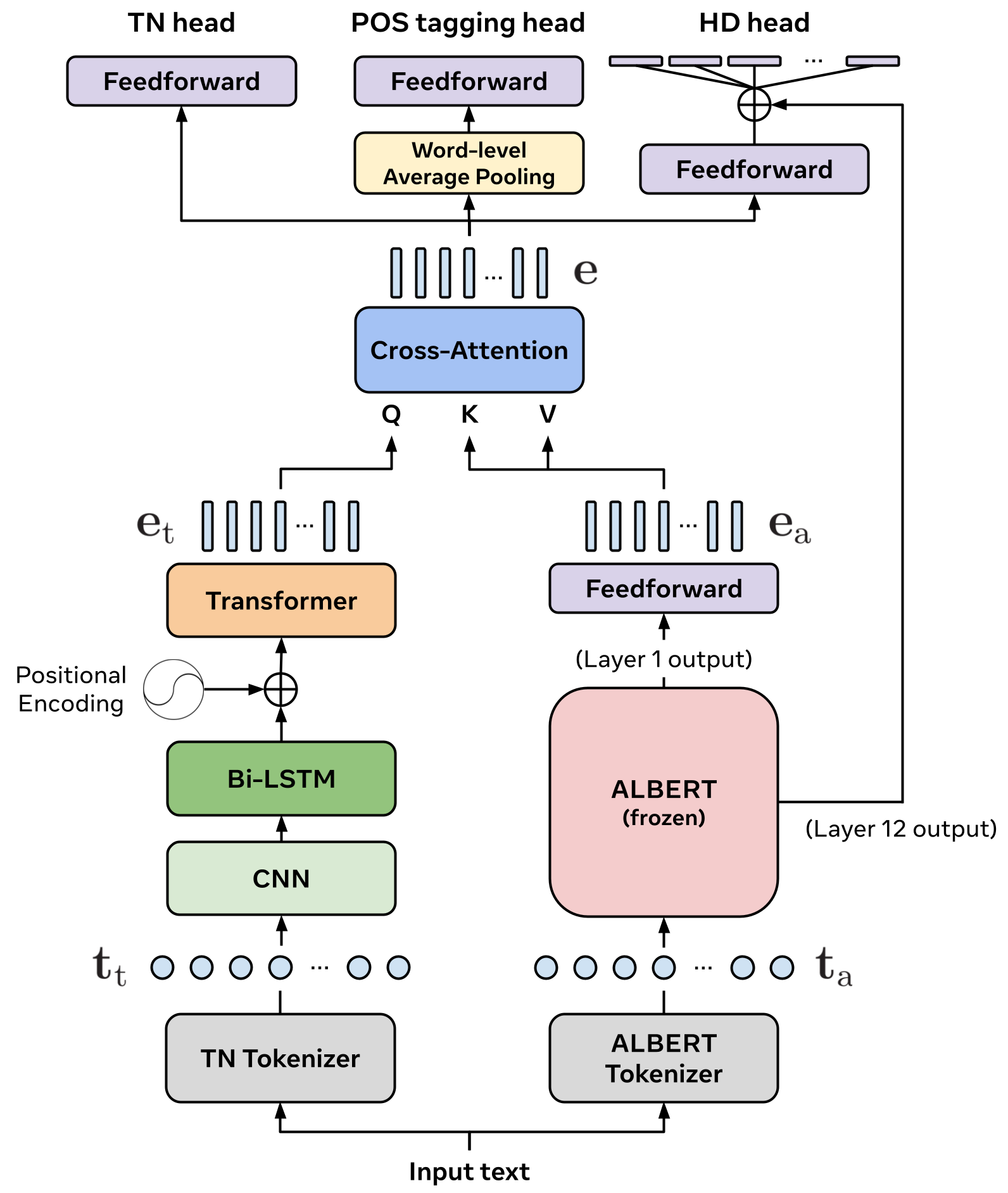}
    \vspace{-4pt}
    \caption{Block diagram of the proposed multi-task model for TN, POS tagging, and HD. The shared trunk processes the input text in two streams, which are combined using cross-attention. The shared representations are then passed to separate heads that solve each task.}
    \label{fig:model}
    \vspace{-13pt}
\end{figure}

\vspace{-18pt}
\section{Proposed Method}
\label{sec:method}

\vspace{-8pt}
\subsection{Preliminaries}
\label{subsec:preliminaries}
\vspace{-5pt}


Our TN system is based on semiotic classification, similar to the one in~\cite{tyagi2021proteno}.
It treats TN as a sequence tagging problem, where input text is split into tokens and the objective is to predict an appropriate rule for normalizing each token.
The TN tokenizer is deterministic; it first splits text on spaces and then further splits it wherever there is a change in the unicode class (e.g., `1/2023' is split into [`1', `/', `2023']).
We use an internally developed token-to-normalization ruleset for American English consisting of 106 rules for 14 semiotic classes. Each rule verbalizes one or more consecutive tokens at a time, and the objective is to solve a 106-way classification problem for each token.
Rules that cannot parse a given token and its successors are masked from the output of the model's final classification layer.
Based on the predictions, a beam search is applied to find the optimal sequence of TN rules to produce the final normalization.

For POS tagging, we use a set of 15 classes: adjective, adverb, article, auxiliary, conjunction, interjection, name, noun, participle, particle, preposition, pronoun, punctuation, spelling, and verb.
Therefore, the task is to solve a 15-way classification problem for each word in a given input sentence.

For HD, we consider the 162 American English homographs from the Wikipedia dataset \cite{gorman2018improving}.
160 of these have two pronunciations, and two have three pronunciations.
We treat HD as a pronunciation classification task; given an input sentence containing a homograph, we predict which pronunciation to use given the surrounding context.

\vspace{-7pt}
\subsection{Model}
\label{subsec:model}
\vspace{-4pt}

Our multi-task model has a tree-like structure with a trunk for shared feature extraction and separate task-specific heads; Figure \ref{fig:model} shows a block diagram of the overall architecture.

\vspace{2pt}
\noindent \textbf{Trunk.}
The trunk takes a piece of text as input and processes it in two information streams.
The first stream operates on a TN token sequence of length $n$ (from the tokenizer described in Section~\ref{subsec:preliminaries}), which we denote as $\mathbf{t}_{\mathrm{t}} = [t_{\mathrm{t}}^{(1)}, t_{\mathrm{t}}^{(2)}, ..., t_{\mathrm{t}}^{(n)}]$.
First, a stack of stateful convolutional layers is applied to each token at the character-level to obtain character embeddings, which are mean pooled to obtain token-level embeddings.
Then, the token embedding sequence is passed through a bidirectional long short-term memory (Bi-LSTM) layer and a Transformer layer~\cite{vaswani2017attention} in order to induce context-sensitivity.
We denote the resulting embedding sequence $\mathbf{e}_{\mathrm{t}} = [e_{\mathrm{t}}^{(1)}, e_{\mathrm{t}}^{(2)}, ..., e_{\mathrm{t}}^{(n)}]$.

The second stream feeds the input text through a pre-trained Transformer-based LM, ALBERT~\cite{lan2019albert}.
We chose to use ALBERT because of its relatively compact size and good performance on various NLP benchmarks; however, it could feasibly be replaced with any other similar LM.
By incorporating this module, our aim is to utilize the additional linguistic knowledge encoded within the LM's embeddings in order to improve performance on our three tasks.
ALBERT operates on a token sequence $\mathbf{t}_{\mathrm{a}}$ that in general has a different length from the TN token sequence; we denote it as having length $m$: $\mathbf{t}_{\mathrm{a}} = [t_{\mathrm{a}}^{(1)}, t_{\mathrm{a}}^{(2)}, ..., t_{\mathrm{a}}^{(m)}]$.
The LM then produces a corresponding embedding sequence $\mathbf{e}_{\mathrm{a}} = [e_{\mathrm{a}}^{(1)}, e_{\mathrm{a}}^{(2)}, ..., e_{\mathrm{a}}^{(m)}]$.

The last part of the trunk combines the two embedding sequences using a cross-attention mechanism.
This module largely follows the structure of a Transformer layer, but instead of self-attention, it uses $\mathbf{e}_{\mathrm{t}}$ as the query and $\mathbf{e}_{\mathrm{a}}$ as the key and value.
Applying cross-attention in this way allows us to combine the two sequences while maintaining the length of $\mathbf{e}_{\mathrm{t}}$ in the output sequence, which we denote as $\mathbf{e} = [e^{(1)}, e^{(2)}, ..., e^{(n)}]$.
This is a desirable design choice because of our TN framework; since rule classification is done at the TN token-level, it is convenient to have an embedding sequence that has a direct one-to-one mapping to the original TN tokens so that we can simply predict a rule for each embedding.

\vspace{2pt}
\noindent \textbf{Task-specific heads.}
The output embedding sequence from the trunk, $\mathbf{e}$, is fed into task-specific heads for TN, POS tagging, and HD.
Each head contains a feedforward module made up of a linear layer and a ReLU activation.
For TN, each embedding in $\mathbf{e}$ is fed through the feedforward module, followed by a final linear layer for token-level rule classification.
For POS tagging, the embeddings in $\mathbf{e}$ are first aggregated into the \textit{word}-level by averaging any token embeddings that make up a single word.
Then, each word-level embedding is fed to the feedforward module and a final linear layer for POS tag prediction.
In addition to the feedforward module, the HD head contains 162 dedicated pronunciation classification heads for each homograph.
The embedding in $\mathbf{e}$ at the index corresponding to the homograph is fed through the feedforward module and the appropriate homograph classification head to predict the pronunciation.

\vspace{2pt}
\noindent \textbf{Incorporating ALBERT effectively.}
Prior work analyzing intermediate representations of Transformer LMs found that layers at various depths learn different structural information about language~\cite{jawahar2019does}.
We performed experiments to determine the optimal layers of ALBERT to use embeddings from.
We found that embeddings from earlier layers (containing syntactical information) were more beneficial for TN and POS tagging, while those from later layers (containing contextual information) were more beneficial for HD; depending on the task, we observed up to a 2\% difference in downstream accuracy between the best and worst layers.
Based on these results, we incorporate ALBERT embeddings in two different ways.
First, we use embeddings from the first layer as inputs to the trunk's cross-attention module in order to influence both TN and POS tagging.
Second, we incorporate embeddings from the final (12th) layer directly into the HD head by taking the embeddings at the indices that correspond to the homograph, aggregating them via averaging, and using a skip connection to add them before the appropriate homograph's classification head.

\vspace{-8pt}
\subsection{Training}
\label{subsec:training}
\vspace{-4pt}

We use cross-entropy loss as the objective function for all three tasks.
To train our model, we cycle through the tasks and perform optimization for only one task within each minibatch.
This is because we use separate datasets for each task, and a given sample can only be used for training on the task that it has labels for.
Therefore, the model is trained for an equal number of iterations on each task; the trunk is optimized with respect to all three tasks over the course of training, while each task-specific head is optimized only on its corresponding task.
We did not perform any kind of task-wise loss weighting or balancing.
While we considered other strategies that stochastically sample tasks or weight losses based on task importance or dataset size (e.g., \cite{luong2016multi}), we found that our method was sufficient for stable convergence and good performance on all three tasks.
\vspace{-6pt}
\section{Llama 2-Generated Homograph Dataset}
\label{sec:hd_dataset}
\vspace{-6pt}

Many recent approaches for HD have utilized the dataset introduced in \cite{gorman2018improving}, which consists of 162 English homographs and 100 sentences per homograph taken from Wikipedia.
However, many of these homographs have a heavily imbalanced number of sentences for each pronunciation, which was also noted in previous work~\cite{nicolis2021homograph}.
For example, of the 90 instances of “abstract” in the training set, 89 are pronounced /\textipa{"\ae b""st\*r\ae kt}/ while only one is pronounced /\textipa{\ae b"st\*r\ae kt}/; in the evaluation set, all 10 instances are pronounced /\textipa{"\ae b""st\*r\ae kt}/.
We argue that such data does not provide enough information for a model to learn to truly disambiguate between pronunciations, nor can it accurately measure a model's capabilities.

To solve this issue, we introduce a new HD dataset encompassing the same 162 English homographs as above, but with an equal number of sentences for each pronunciation of each word, generated using Llama 2-Chat 70B~\cite{touvron2023llama2}.
In creating this dataset, we aimed to follow two principles.
First, the dataset must be \textit{balanced}: we wanted an equal number of sentences for each pronunciation of each homograph, which should be stratified evenly across train and test sets.
Second, the dataset must be \textit{diverse}: for each pronunciation of each homograph, as many word senses as possible should be captured, and the word should appear in as many domains and play as many different roles in a sentence as possible.
The resulting dataset contains 10 sentences per pronunciation per homograph, for a total of 3,260 sentences.
\begin{table*}[t]
    \centering
    \caption{TN, POS tagging, and HD evaluation results. We show results for the full model trained on all three tasks, as well as versions with ablated components or that were trained only on individual or sub-combinations of tasks.
    }
    \vspace{-5pt}
    \footnotesize
    \begin{tabular}{lccccccc}
    \toprule
    \multirow{3}{*}{\vspace{-5pt} Model} &
      \multicolumn{2}{c}{\textbf{TN}} &
      \multicolumn{2}{c}{\textbf{POS Tagging}} &
      \multicolumn{3}{c}{\textbf{HD}} \\
      \cmidrule(lr){2-3}
      \cmidrule(lr){4-5}
      \cmidrule(lr){6-8}
    \multicolumn{1}{c}{} &
      \makecell{Line Accuracy} &
      WER &
      \makecell{Accuracy \\ (SwDA)} &
      \makecell{Accuracy \\ (Internal)} &
      \makecell{Micro \\ (Wikipedia)} &
      \makecell{Macro \\ (Wikipedia)} &
      \makecell{Micro/Macro \\ (Llama 2)} \\ \midrule
    Proposed (TN + POS + HD)             & \textbf{86.93} & 2.40           & 97.18          & 89.91            & \textbf{96.84}  & \textbf{93.10}   & 93.56 \\
    \quad -- residual connection for HD  & 86.00          & 2.64           & 97.18          & 90.38            & 96.59  & 92.11            & \textbf{94.79} \\
    \quad\quad -- ALBERT                 & 84.00          & 2.84           & 96.12          & 87.21            & 96.28  & 91.86            & 92.58          \\ \midrule
    TN + POS                             & 85.07          & 2.70           & 97.54          & 90.98            & --     & --               & --             \\
    TN + HD                              & 85.47          & 2.87           & --             & --               & 95.42  & 89.70            & 88.77          \\
    POS + HD                             & --             & --             & 97.19          & 90.11            & 96.72  & 92.40            & 93.56          \\ \midrule
    TN only                              & 86.53          & \textbf{2.38}  & --             & --               & --     & --               & --             \\
    POS only                             & --             & --             & \textbf{97.58} & \textbf{91.30}   & --     & --               & --             \\
    HD only                              & --             & --             & --             & --               & 93.93  & 86.92            & 87.48          \\
    \bottomrule
    \end{tabular}
    \label{table:main_results}
    \vspace{-6pt}
\end{table*}

\vspace{-5pt}
\section{Experiments}
\vspace{-9pt}

\subsection{Configurations}
\label{subsec:config}

\vspace{-4pt}
\noindent \textbf{Model parameters.}
For the TN input stream in the trunk of our model, we used character embeddings of size 32.
We used 1 stateful convolutional layer with a channel size of 64, kernel size of 5, dropout with $p = 0.2$, and batch normalization~\cite{ioffe2015batch} with a ReLU activation.
The Bi-LSTM used a hidden size of 128, resulting in an output hidden state of size 256.
For the Transformer and cross-attention modules, we set the hidden sizes to 256, the number of attention heads to 4, and used dropout with $p = 0.1$.
Feedforward modules in each task-specific head used linear layers of size 256.

\vspace{2pt}
\noindent \textbf{Data.}
For TN, we used an internal dataset consisting of 37k training, 2k validation, and 750 test sentences.
For POS tagging, we used the Switchboard Dialog Act (SwDA) Corpus (125k sentences)~\cite{jurafsky1997switchboard} and an internal dataset of sample responses from a speech assistant (1k sentences).
The original POS tags in the SwDA Corpus were condensed into the 15 classes described in Section \ref{subsec:preliminaries}.
We held out 0.5\% of the SwDA Corpus each for validation and testing and used the internal dataset for testing only, for a total of 124k training, 627 validation, and 1.6k test samples.
For HD, we used the Wikipedia dataset from \cite{gorman2018improving} and the Llama 2 dataset described in Section \ref{sec:hd_dataset}.
We held out 10\% of the Wikipedia training set for validation and used its evaluation set as is for testing.
For the Llama 2 dataset, we evenly split the sentences into train and test sets, stratified by homograph pronunciations, but did not hold out a validation set in order to maximize usage of its sentences during training.
This made for a total of 15k training, 1.5k validation, and 3.2k test samples.

\vspace{2pt}
\noindent \textbf{Training.}
All experiments were conducted on a single NVIDIA A100 GPU.
We trained our model for 90k iterations (30k iterations per task) using the AdamW optimizer~\cite{loshchilov2018decoupled} with learning rate 5e-4 and $\beta_1 = 0.9, \beta_2 = 0.99$.
The batch size was set to 128, and the learning rate was decayed to 20\% of its value every 16k steps.
ALBERT weights were kept frozen throughout the course of training.

\vspace{2pt}
\noindent \textbf{Evaluation.}
We evaluated TN performance using line accuracy (whether the predicted normalization exactly matches the ground truth) and word error rate (WER).
For POS tagging, we evaluated using accuracy, and for HD, we used both micro- and macro-average accuracy over the homograph pronunciation classes.

\vspace{-7pt}
\subsection{Results}
\label{subsec:results}
\vspace{-4pt}

We compare our full multi-task model trained on all three tasks against task-ablated versions trained only on individual or combinations of two out of three tasks.
The task-ablated models have the same architecture as the full model except for the absence of task-specific heads for removed tasks, and were trained for 30k iterations per task (30k iterations for single-task models and 60k iterations for two-task models).
The results are shown in Table \ref{table:main_results}.
Note that for HD, micro- and macro-average accuracies on the Llama 2 dataset are identical because all homograph classes have the same number of samples.

Overall, we find that multi-task learning has clear benefits for performance.
When comparing two-task models against single-task models, we find that HD performance improves significantly when trained together with either TN or POS tagging.
TN performance drops somewhat when trained together with an additional task, and POS tagging performance also drops marginally.
However, our full model trained on all three tasks achieves the strongest performance overall, improving upon or matching the performance of all single- or two-task models on TN and HD.
POS tagging performance drops slightly compared to the single-task POS tagging model; this mirrors the results in \cite{changpinyo2018multi}, which found POS tagging to be beneficial to other tasks but also often harmed by them in MTL.
This could be because POS tagging is a simpler problem than TN or HD that does not require as much contextual information to solve.
However, given that the performance differences are small, and that TN and HD carry more practical importance in a TTS front-end, we consider this minor drop-off to be relatively inconsequential.
We also verified that any differences were not simply due to multi-task models being trained longer, as we did not find any further performance improvements from training single- or two-task models for more iterations.
Overall, these results point to the presence of meaningful positive transfer between the three tasks, validating our hypothesis for the MTL framework.

\vspace{-8pt}
\subsection{Ablation studies}
\label{subsec:ablations}
\vspace{-4pt}

We conducted ablation studies on key components of our model; the results are shown in the top section of Table \ref{table:main_results}.
When the residual connection from the final layer of ALBERT to the HD head is removed, HD performance decreases on the Wikipedia dataset, but improves slightly on the Llama 2 dataset.
While this indicates that the contribution of ALBERT's final layer embeddings towards HD is inconclusive, it shows that they can have a positive impact depending on the setting; more in-depth studies may be needed on how to most effectively incorporate them into the model.
Alternatively, this opens up the possibility of pruning ALBERT's weights except for the first layer, which would make the overall model significantly smaller (at the cost of a slight drop in TN performance).
When ALBERT is removed from the model altogether, performance on TN and POS tagging further drop significantly, demonstrating that the syntactical information in its first layer's embeddings is crucial for those tasks.

\subsection{Impact of the Llama 2 homograph dataset}
\label{subsec:llama2_hd_results}
\vspace{-5pt}

We analyzed the impact of our proposed Llama 2 homograph dataset on HD performance.
To do this, we compared versions of our full multi-task model trained on either only the Wikipedia dataset or both the Wikipedia and Llama 2 datasets; the TN and POS datasets were kept constant.
We did not experiment with training on the Llama 2 dataset alone due to its relatively small size.

Table \ref{table:hd_data_results} shows the micro and macro homograph prediction accuracies on the two test sets.
For brevity, we do not show results on the other two tasks because we did not find significant differences.
We see that training on both datasets yields higher accuracies compared to training on only the Wikipedia dataset.
There are small performance gains on the Wikipedia test set, with a slight improvement in macro-average accuracy.
However, the most significant improvements come on the Llama 2 test set, with absolute accuracy improvements of around 9\%.
Notably, the Wikipedia-only model exhibits a large performance gap between the two test sets, while the model trained on both datasets achieves similar performance on both.
In addition, for both models, there is a gap between the micro- and macro-average accuracies on the Wikipedia test set, while the values are identical on the Llama 2 test set; this reflects the balance (or lack thereof) of homograph classes in each test set.

\begin{table}[t]
    \centering
    \caption{HD accuracies of the full multi-task model when trained on both Wikipedia and Llama 2 datasets vs. only the Wikipedia dataset.}
    \vspace{-5pt}
    \footnotesize
    \begin{tabular}{lclc}
    \toprule
    \multirow{2}{*}{\vspace{-5pt} HD Training Data} &
    \multicolumn{2}{c}{Wikipedia}  &
    Llama 2                        \\
    \cmidrule(lr){2-3}
    \cmidrule(lr){4-4}
    & \multicolumn{1}{l}{Micro} 
    & \multicolumn{1}{l}{Macro} 
    & Micro/Macro   \\ \midrule
    Wikipedia + Llama 2 & \textbf{96.84} & \textbf{93.10} & \textbf{93.56} \\
    Wikipedia-only      & \textbf{96.84} & 92.04          & 84.54          \\ \bottomrule
    \end{tabular}
    \vspace{-11pt}
    \label{table:hd_data_results}
\end{table}
\vspace{-5.5pt}
\section{Conclusion}
\label{sec:conclusion}
\vspace{-5.5pt}

In this paper, we proposed a multi-task model that jointly learns to solve three tasks that are common components of a text-to-speech (TTS) front-end: text normalization (TN), part-of-speech (POS) tagging, and homograph disambiguation (HD).
We demonstrated the benefits of multi-task learning in this setting, showing that our full model trained on all three tasks achieves the strongest overall performance compared to models trained on individual or sub-combinations of two tasks.
In addition, we introduced a new HD dataset that contains balanced and diverse sentences for each pronunciation of 162 American English homographs, and showed that it significantly helps with improving and more accurately measuring HD performance.
These findings may provide valuable insights for future work on developing more unified TTS front-ends.

\clearpage
\bibliographystyle{IEEEbib}
\bibliography{refs}

\end{document}